 \documentclass[tablecaption=bottom,pmlr]{jmlr}

\usepackage{booktabs}
\usepackage[load-configurations=version-1]{siunitx} % newer version
\theorembodyfont{\upshape}
\theoremheaderfont{\scshape}
\theorempostheader{:}
\theoremsep{\newline}

\usepackage{algorithm}
\usepackage{algorithmicx}
\usepackage{algpseudocode}
\usepackage{mathtools}
\newcommand{\eg}{\textit{e}.\textit{g}. }
\usepackage{xcolor}
\usepackage{multirow}

\newcommand\B{\rule[-1.2ex]{0pt}{0pt}} % Bottom strut

\jmlrvolume{148} 
\jmlryear{2021}
\jmlrsubmitted{submission date}
\jmlrpublished{publication date}
\jmlrworkshop{NeurIPS 2020 Preregistration Workshop} % W&CP title

\title[SFTrack++: A Spectral Segmentation Approach for Space-Time Consistent Tracking]{SFTrack++: A Fast Learnable Spectral Segmentation Approach for Space-Time Consistent Tracking}

\firstpageno{136}

\author{\Name{Elena Burceanu} \Email{eburceanu@bitdefender.com}\\
 \addr  Bitdefender\\
  University of Bucharest, Romania\\
  Institute of Mathematics of the Romanian Academy\\}

\begin{document}

\maketitle

\begin{abstract}
We propose an object tracking method, SFTrack++, that smoothly learns to preserve the tracked object consistency over space and time dimensions by taking a spectral clustering approach over the graph of pixels from the video, using a fast 3D filtering formulation for finding the principal eigenvector of this graph's adjacency matrix. To better capture complex aspects of the tracked object, we enrich our formulation to multi-channel inputs, which permit different points of view for the same input. The channel inputs are in our experiments, the output of multiple tracking methods. After combining them, instead of relying only on hidden layers representations to predict a good tracking bounding box, we explicitly learn an intermediate, more refined one, namely the segmentation map of the tracked object. This prevents the rough common bounding box approach to introduce noise and distractors in the learning process. We test our method, SFTrack++, on five tracking benchmarks: OTB, UAV, NFS, GOT-10k, and TrackingNet, using five top trackers as input. Our experimental results validate the pre-registered hypothesis. We obtain consistent and robust results, competitive on the three traditional benchmarks (OTB, UAV, NFS) and significantly on top of others (by over $1.1\%$ on accuracy) on GOT-10k and TrackingNet, which are newer, larger, and more varied datasets. The code is available at \url{https://github.com/bit-ml/sftrackpp}.
\end{abstract}

\section{Introduction}

% better temporal use
Better using the temporal aspect of videos in visual tasks has been actively discussed for a rather long time, especially with the large and continuous progress in hardware. The first aspect we tackle in our approach is a seamless blending of space and time dimensions in visual object tracking.
Current methods mostly rely on target appearance and frame-by-frame processing~\cite{osvos, siamrpn++, atom}, with rather few taking explicit care of temporal consistency~\cite{kys, cycle-random-walk}. In the spectral graph approach, nodes are pixels and edges are their local relations in space and time, while the strongest cluster in this graph, given by the principal eigenvector of the graph's adjacency matrix, represents the consistent main object volume over space and time.

\begin{figure*}[t]
	\begin{center}
		\includegraphics[width=0.99\linewidth]{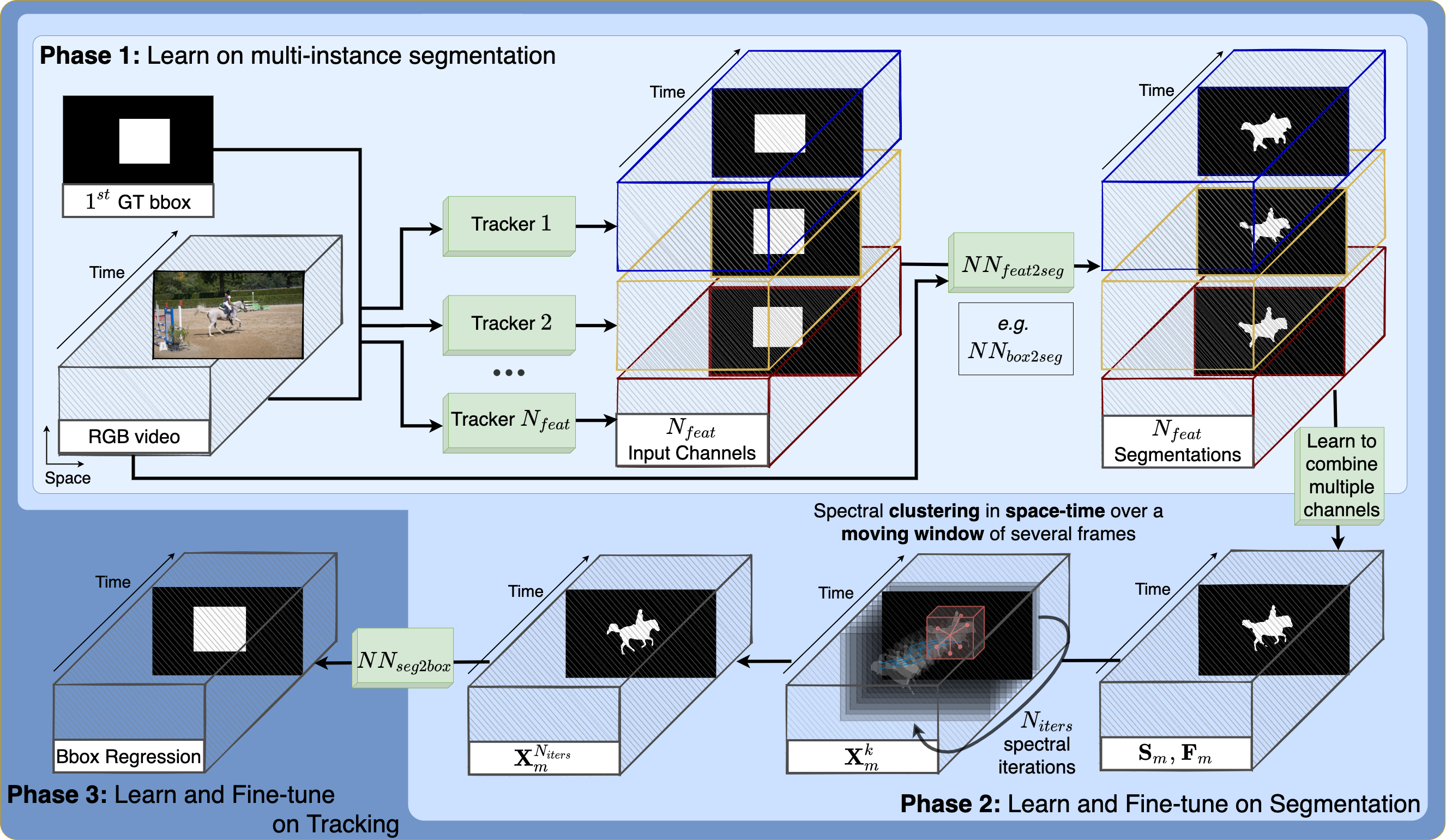}
	\end{center}
	\caption{SFTrack++: We start from video's RGB and $1^{st}$ frame GT bbox of the tracked object. We run state-of-the-art trackers, in an online manner, while fine-tuning $NN_{feat2seg}$ network frame-by-frame (pretrained in Phase 1) to transform the extracted feature maps (\eg bboxes) to segmentation maps. Next, we learn to combine multiple segmentation inputs and refine  the final mask using a spectral approach, applied also online over a moving window containing the previous N frames, for $N_{iter}$ spectral iterations (Phase 2). In Phase 3, we learn a bbox regressor from the final segmentation mask, $NN_{seg2box}$ and fine-tune all our parameters on the tracking task.}
	\label{fig: architecture}
\end{figure*}

% bbox vs segmentation
A second observation challenges the rough bounding box (bbox) shape used for tracking. While it provides a handy way to annotate datasets, it is a rather imperfect label since it leads to errors that accumulate, propagate, and are amplified over time. Objects rarely look like boxes, and bboxes contain most of the time significant background information or distractors. Since having a good segmentation for the interest object directly influences the tracking performance, we constrain an intermediary representation, a segmentation map, which aims to reduce the quantity of noise transferred from a frame to the next one. We integrate it into our end-to-end flow, as shown in Fig.~\ref{fig: architecture}.

% ensemble
A third point we emphasize on is relying on multiple, independent characteristics of the same object or multiple modules specialized in different aspects. This comes with an improved ability to understand complex objects while increasing the robustness~\cite{kys}. 
We, therefore, adjust the SFSeg spectral approach, enhancing its formulation to support learning on top of multiple input channels. They could be general features for the input frame coming from different approaches or, more specifically, tracking outputs from multiple solutions, as we test in Sec.~\ref{sec: experiments}.

The \textbf{main contributions} of our approach are:
\begin{itemize}
    \item SFTrack++ brings to tracking a natural, contiguous, and efficient approach for integrating space and time components, using a fast 3D spectral clustering method over the graph of pixels from the video, to strengthen the tracked object's model.
    \item We explicitly learn intermediate fine-grained segmentation as opposed to rough bounding boxes in our three phases end-to-end approach to a more robust tracking solution.
    \item We integrate into our formulation a way of learning to combine multiple input channels, offering a wider view of the objects, harmonizing different perceptions, for a powerful and robust approach.
\end{itemize}

\section{Related work}
\noindent\textbf{General Object Tracking.} Out of the three main trackers families, \textbf{Siamese based trackers} gained a lot of traction in recent years for their high speed and end-to-end capabilities \cite{siamfc, siam-rpn, siamrpn++, ocean}. Most approaches focus on exhaustive offline-training, failing to monitor changes w.r.t. the initial template \cite{siam-inst-search, siamfc++, siamban}, while others update their model online ~\cite{siam-rcnn, ltmu}. Nevertheless, the robustness towards unseen objects and transformations at training time remains a fundamental problem for Siamese trackers. \textbf{Meta-learning approaches for tracking}~\cite{meta-tracker, oneshot-tracker, meta-learning-tracker} come with an interesting way of adapting to the current object of interest, while keeping a short inference time, by proposing a target-independent tracking model.
One major limitation for both those approaches, Siamese and meta-learning trackers, is that they fail to adapt continuously to the real-time changes in the tracked object, using rather a history of several well-chosen patches or even just the initial one. In contrast, our method naturally integrates the temporal dimension, by continuously enforcing the local temporal and spatial object consistency. \textbf{Discriminative methods}~\cite{atom, prdimp, d3s} on the other hand are classic approaches, focusing more on changes in the tracked object \cite{eco} (background, distractors, hard negatives), better integrating the temporal dimension \cite{kys} in the method flow. They prove to be robust, but they mostly rely on hand-crafted observations or modules not trainable end-to-end. SFTrack++ provides an end-to-end approach while minimizing the distractors and background noise using the intermediary segmentation map. Our method distances itself from a certain family of trackers, introducing the space and time consistency endorsement via clustering component as an additional dimension of the algorithm.

% bbox vs segmentare; unique feature extractor vs multi-channel
With a few notable exceptions \cite{segtrack, mots}, most tracking solutions use internally hidden layer representations extracted from the previous frame's rough bbox prediction \cite{kys, prdimp, atom, siamfc++}, rather than a fine-grained segmentation mask as in our approach. Also, most of them do not take into account multiple perceptions for the input frame and operate over a unique feature extractor~\cite{atom, ocean, siamfc}. There are a few trackers though that combine two models for adapting to sudden changes while remaining robust to background noise, by explicitly model the different pathways \cite{kys, multi-path-tracking}. In contrast, our end-to-end multi-channel formulation learns over $10$ input channels, a significantly larger number.

%: a more detailed representations are costly, but achieves better results, while graphs with fewer nodes are faster but neglect particularities. 
\noindent\textbf{Graph representations.} Images and videos were previously represented as graphs, where the nodes are pixels, super-pixels, or regions \cite{contrastive-random-walk}. This choice directly impacts the running time and performance. Regarding edges, they are usually undirected, modeled by symmetric similarity functions, but there are also several works that use directed ones \cite{directed-affinities, directed-groups}. \textbf{Spectral clustering} approaches~\cite{NJW, meila_shi, marius_iccv2005} search for the leading or the smallest eigenvector for the graph's adjacency matrix to solve the clusters' assignments. Spectral clustering was previously used in pixel-level image segmentation~\cite{img_normalized_cut_malik_2000}, with a high burden on the running time and in building space-time correspondences between video patches~\cite{contrastive-random-walk}. Graph Cuts is a common approach for spectral clustering, having many variations \cite{img_normalized_cut_malik_2000, cut_min_max, cut_average}. SFSeg~\cite{sfseg} proposes a 3D filtering technique for efficiently finding the spectral clustering solution without explicitly computing the graph's adjacency matrix. Inspired by this method, we integrate an improved version with learning over multi-channel inputs, as an intermediate component in our tracker, as detailed in Sec.~\ref{sec: method}.

\section{Our approach}
\label{sec: method}

SFTrack++ algorithm has three phases, as we visually present them in Fig.~\ref{fig: architecture}. In \textbf{Phase~1}, we learn a neural net, $NN_{feat2seg}$, that transforms the RGB and a frame-level feature map extracted using a tracker (\eg bbox from a tracker prediction) into a segmentation mask. Using only the RGB as input is not enough, because frames can contain multiple objects and instances, and we also need a pointer to the tracked object to predict its segmentation. Next, in \textbf{Phase~2}, we run multiple state-of-the-art trackers frame-by-frame over the input as an online process and extract input channels from them (\eg bboxes). We transform those feature maps to segmentation maps with the previously recalled module, $NN_{feat2seg}$. Next, we learn to combine and refine the outputs for the current frame using a spectral solution for preserving space-time consistency, adapted to learn over multiple channels. Note that, when applying the spectral iterations, we use a sliding window approach over the previous N frames in the video volume. For supervising this path, we use segmentation ground-truth. \textbf{Phase~3} learns a neural net as a bbox regressor over the final segmentation map from the previous phase, $NN_{seg2box}$, while fine-tuning all the other trainable parameters in the model, using tracking GT.

\noindent\textbf{Spectral approach to segmentation.} We go next through the following aspects, briefly explaining the connection between them: segmentation $\rightarrow$ leading eigenvector $\rightarrow$ power iteration $\rightarrow$ 3D filtering formulation $\rightarrow$ multi-channel. Image segmentation was previously formulated as a graph partitioning problem, where the segmentation solution \cite{img_normalized_cut_malik_2000} is the leading eigenvector of the adjacency matrix. It was used in a similar way for video \cite{sfseg}. Power iteration algorithm can compute the leading eigenvector: $\mathbf{x}^{k+1}_i \leftarrow \sum_{j \in \mathcal{N}(i)}\mathbf{M}_{i, j} \mathbf{x}^k_j$, where $\mathbf{M}$ is the $N \times N$ graph's adjacency matrix, $N$ is the number of nodes in the graph (pixels in the video space-time volume in our case), $\mathcal{N}(i)$ is the space-time neighbourhood of node $i$ and each step $k$ is followed by normalization. The adjacency matrix used in power iteration usually depends on two types of terms: unary ones are about individual node properties and pairwise ones describe relations between two nodes (pairs). 

Following this approach, SFSeg rewrites power iteration using 3D filtering for an approximated adjacency matrix. The solution is described in Eq.~\ref{eq: sfseg++_1}:
\begin{equation}
    \begin{aligned} 
        \mathbf{X}^{k+1} \leftarrow \mbox{normalized}(\mathbf{S}^p \cdot (\alpha^{-1} \mathbf{1} - \mathbf{F}^2) \cdot G_{3D} * (\mathbf{S}^p  \cdot \mathbf{X}^k) -
                    \mathbf{S}^p \cdot G_{3D} * (\mathbf{F}^2 \cdot \mathbf{S}^p \cdot \mathbf{X}^k) +\\
                     2 \mathbf{S}^p \cdot \mathbf{F} \cdot G_{3D} * (\mathbf{F} \cdot \mathbf{S}^p \cdot \mathbf{X}^k)),
    \end{aligned}
\label{eq: sfseg++_1}
\end{equation}
where $*$ is a 3D convolution with Gaussian filter $G_{3D}$ over space-time volume, $\cdot$ is an element-wise multiplication, $\mathbf{S}$ and $\mathbf{F}$ are unary and pairwise terms in matrix form with $p$ and $\alpha$ controlling their importance, $k$ is the current spectral iteration and $\mathbf{X}, \mathbf{S}, \mathbf{F}$ matrices have the original video shape ($N_{frames}\times H \times W$).

\noindent\textbf{Multi-channel learning formulation.} We extend the single-channel formulation in SFSeg such that it can learn how to combine several input channels, $S_i$ and $F_i$, for unary and pairwise terms respectively: $\mathbf{S}_m \leftarrow \sigma(\sum_{i=1}^{N_{cs}} w_{s, i} \mathbf{S}_i + b_s \mathbf{1})$, $\mathbf{F}_m \leftarrow  \sigma(\sum_{i=1}^{N_{cf}} w_{f, i} \mathbf{F}_i  + b_f \mathbf{1})$, where $\mathbf{S}_m$ and $\mathbf{F}_m$ are the multi-channel unary and pairwise maps, respectively, $\sigma$ is the sigmoid function, $N_{cs}$ and $N_{cf}$ are the number of input channels, $\mathbf{1}$ is an all-one matrix for the bias terms and $w_{s, i}, w_{f, i}, b_s, b_f$ are their corresponding learnable weights. We replace $\mathbf{S}$ and $\mathbf{F}$ in Eq.~\ref{eq: sfseg++_1} with their multi-channel versions $\mathbf{S}_m$ and $\mathbf{F}_m$, respectively. We learn $w_{s, i}, w_{f, i}, b_s, b_f$ parameters both over segmentation and tracking tasks. More, SFTrack++ can learn end-to-end, from the original input frames all the way to final output, in the case of end-to-end learnable feature extractors.

\section{Experimental protocol}
\label{sec: experiments}
We test if SFTrack++ brings in a complementary dimension to tracking by having an intermediary fine-grained representation, extracted over multiple state-of-the-art trackers' outputs, and smoothed in space and time. We guide our experiments such that we evaluate the least expensive pathways first. For reducing the hyper-parameters search burden, we use AdamW~\cite{adamw-amsgrad}, with a scheduler policy that reduces the learning rate on a plateau. For efficiency and compactness, we use the same channels to construct both the unary and pairwise maps: $\mathbf{S}_i = \mathbf{F}_i$. Their learned weights are also shared $w_{s, i} = w_{f, i}$. We use as input channels bboxes extracted with top single object trackers. We choose 10 top trackers: SiamR-CNN~\cite{siam-rcnn}, LTMU~\cite{ltmu}, KYS~\cite{kys}, PrDiMP~\cite{prdimp}, ATOM~\cite{atom}, Ocean~\cite{ocean}, D3S~\cite{d3s}, SiamFC++~\cite{siamfc++}, SiamRPN++~\cite{siamrpn++}, SiamBAN~\cite{siamban}, which differ in architecture, training sets and overall in their approaches, but all achieves top results on tracking benchmarks.

\noindent\textbf{Training.} In Phase 1 we train our $NN_{feat2seg}$ network on DAVIS-2017~\cite{davis-2017} and Youtube-VIS~\cite{Youtube-VIS} trainsets, for each individual object. It receives the current RGB and the output of a tracking method (bbox), randomly sampled at training time. We use the U-Net architecture, validating the right number of parameters ($100$K - $1$ mil) and the number of layers. We use DAVIS-2017 and Youtube-VIS evaluation sets to stop the training. Following the curriculum learning approach, before introducing tracking methods intro the pipeline, we use at the beginning of the tracking GT bboxes (extracted from segmentation GT, as straight bboxes). This allows the $NN_{feat2seg}$ component to get a good initialization, before introducing faulty bbox extractors, namely the top 10 tracking methods mentioned before. For Phase 2, we also train for the segmentation task. We learn the second part of our method to have an intermediary fine-grained representation, extracted over multiple channels, and smoothed in space and time. We validate here $N_{iters}$, the number of spectral iterations ($1$-$5$). We train on DAVIS-2016~\cite{davis-2016} and Youtube-VIS datasets. In Phase 3, training for tracking, we learn a regression network, $NN_{seg2box}$ (with $50$K-$500$K parameters), to transform the final segmentation to bbox. We train on TrackingNet~\cite{tracking-net}, LaSOT~\cite{lasot} and GOT-10k~\cite{got-10k} training splits.

\noindent\textbf{Baselines Comparison.}
Experiment for comparing with other methods focus on the improvements SFTrack++ could bring over state-of-the-art and other competitive approaches for general object tracking: single method state-of-the-art solutions, a basic ensemble over the trackers, SFTrack++ applied only over the best tracker, SFTrack++ applied over the basic ensemble and the best learned neural net ensemble we could get out of several configurations (2D and 3D versions for U-Net~\cite{unet} and shallow nets, having a different number of parameters: $100$K, $500$K, $1$ mil, $5$ mil, $15$ mil). All methods receive the same input from top $10$ trackers and train on TrackingNet, LaSOT and GOT-10k train sets as previously described. We evaluate our solution against all baselines on seven tracking benchmarks: \textbf{VOT2018}~\cite{vot-18}, \textbf{LaSOT}, \textbf{TrackingNet}, \textbf{GOT-10k}, \textbf{NFS}~\cite{nfs}, \textbf{OTB-100}~\cite{otb100} and \textbf{UAV123}~\cite{uav123}. For the main conclusion of the paper, we will provide statistical results (mean and variance over several runs) to better indicate a strong positive/negative result, or an inconclusive one.  

\noindent\textbf{Ablative studies.} We vary several components of our end-to-end model to better understand their role and power. We train our \textbf{Phase 1} component, $NN_{feat2seg}$ net, not only for bbox input features but also for other earlier features, extracted from each tracker architecture. We test the overall tracking performance for this case.
We remove from the pipeline the spectral refinement in \textbf{Phase 2} and report the results.
We test the performance of our tracker without the \textbf{Phase 3} neural net, $NN_{seg2box}$, by replacing it with a straight box and rotated box extractors from OpenCV~\cite{opencv}.
We test several losses to optimize for both segmentation and tracking tasks: a linear combination between the weighted diceloss~\cite{wdice-loss} and binary cross-entropy, Focal-Tversky~\cite{focal-tversky}, and Focal-Dice~\cite{focal-dice}. For the ablative experiments, we evaluate only on OTB100, UAV123, and NFS30 tracking datasets.

\section{Experimental Results} 
\label{sec: results}
\begingroup
\setlength{\tabcolsep}{5pt} % Default value: 6pt
\begin{table}[t]
    \begin{center}
        \begin{tabular}{l| r c  c  c ccc  ccc}
        \toprule
        & \multicolumn{1}{c}{\textbf{Method}} &
        \multicolumn{1}{c}{\textbf{OTB}} &
        \multicolumn{1}{c}{\textbf{UAV}} &
        \multicolumn{1}{c}{\textbf{NFS}} &
        \multicolumn{3}{c}{\textbf{GOT-10k}} &
        \multicolumn{3}{c}{\textbf{TrackingNet}} \\
        & & AUC & AUC & AUC & AO & SR$_{50}$  & SR$_{75}$ & Prec & Prec$_{norm}$ & AUC\\
        % \midrule
        \cmidrule(l){1-1}
        \cmidrule(l){2-2}
        \cmidrule(l){3-3}
        \cmidrule(l){4-5}
        \cmidrule(l){6-8}
        \cmidrule(l){9-11}
        \multirow{5}{*}{\rotatebox[origin=c]{90}{Single Method}}
        & D3S  &  57.7 & 45.0 & 38.6 &
        39.3 & 39.0 & 10.1 &
        52.2 & 67.9 & 52.4\\
        & SiamBAN   &  67.6 & 60.8 & 54.2 & 
        54.6 & 64.6 & 40.5 & 
        68.4 & 79.5 & 72.0 \\
        & ATOM-18    &  66.7 & 64.3 & 58.4 & 
        55.0 & 62.6 & 39.6 & 
        64.8 & 77.1 & 70.3\\
        & SiamRPN++ &  65.0 & \textbf{\color{red}65.0} & 50.0 & 
        51.7 & 61.5 & 32.5 & 
        \textbf{\color{red}69.3} & 80.0 & 73.0\\
        & PrDimp-18  &  \textbf{\color{red}67.6} & 63.5 & \textbf{\color{red}62.6} &
        \textbf{\color{red}60.8} & \textbf{\color{red}71.0} & \textbf{\color{red}50.3} & 
        69.1 & \textbf{\color{red}80.3} & \textbf{\color{red}75.0} \B \\
        % & SFTrack++(best)  & 67.1 & 59.7 & 61.3 & 59.8 & 70.4 & \textbf{48.0} & 68.3 & 79.1 & 74.22\\
        \cmidrule(l){1-1}
        \cmidrule(l){2-2}
        \cmidrule(l){3-3}
        \cmidrule(l){4-5}
        \cmidrule(l){6-8}
        \cmidrule(l){9-11}
        \multirow{4}{*}{\rotatebox[origin=c]{90}{Ensemble}}
        & Basic (median) & 66.6 & 60.8 & 55.5 & 
        54.7 & 63.9 & 31.6 &
        69.0 & 80.0 & 73.9\\
        & Neural Net & \textbf{\color{blue}71.3} & 59.7 & 58.2 & 
        59.5 & 69.8 & 42.9 &
        70.6 & 80.2 & 74.5 \\
        & \textbf{SFTrack++} & 70.3 & \textbf{\color{blue}61.2} & \textbf{\color{blue}62.4} & 
        \textbf{\color{blue}62.0} & \textbf{\color{blue}73.3} & \textbf{\color{blue}47.8} & 
        \textbf{\color{blue}71.9} & \textbf{\color{blue}81.9} & \textbf{\color{blue}76.1}\\
        & std & $\pm$ \small 0.5 & $\pm$ \small 0.2 & $\pm$ \small 0.1 & $\pm$ \small 0.7 & $\pm$ \small 0.5 & $\pm$ \small	1.1& $\pm$ \small	0.3 & $\pm$ \small	0.3 & $\pm$ \small	1.0\\
        \bottomrule
        \end{tabular}
    \end{center}
    \caption{Comparison on 5 tracking benchmarks. In the first group we show individual methods, used as input for our SFTrack++. In the second one, we show ensemble methods: a basic (median) and a neural net model with a similar number of parameters like SFTrack++. Our method outperforms both the input or other ensemble methods by a large margin on the challenging benchmarks GOT-10k and TrackingNet, while obtaining competitive results on OTB, UAV, and NFS. For SFTrack++ we report mean and std when training the model from scratch three times. With blue we represent the best single method in the column and with red the best ensemble. The raw results are available in the supplementary material.}
    \label{tab:baselines}
\end{table}
\endgroup

\paragraph{Comparison with other methods.} In Tab.~\ref{tab:baselines} we present the results of our method on five tracking benchmarks: OTB100, UAV123, NFS30, GOT-10k, and TrackingNet, comparing it with other top single methods and ensemble solutions. For single methods, we take into account each input method in our SFTrack++: D3S, SiamBAN, ATOM-18, SiamRPN++, PrDimp-18. We chose only one lightweight configuration per tracker, common across all benchmarks. For ensembles we use 1) a basic per-pixel median ensemble followed by the same bounding box regressor used in all experiments (see Sec.~\ref{sec: modification}) and we also trained a more complex one: 2) a neural net having an UNet architecture (with $5$ down-scaling and $5$ up-scaling layers) and a similar number of parameters like SFTrack++ ($\approx4.3$ millions). We observe that the variation across different runs (including training from scratch) of our method is very small, showing a robust result and a clear conclusion. On newer, larger, and more generic datasets like GOT-10k and TrackingNet, our method surpasses others by a large margin, while on OTB100, UAV123, and NFS30 it has competitive results.

\paragraph{Ablation studies.} To validate the components of our method, we test in Tab.~\ref{tab:ablations} different variations, reporting results on OTB100, UAV123, and NFS30. First, we remove the spectral refining component from phase 2, taking out the temporal dependency and leaving the per frame predictions independent. In the next experiment, we remove the neural net from phase3 $NN_{segm2bbox}$. In the next chunk, we investigate the number of input methods. Last, we vary the number of spectral iterations from phase 2. The conclusions from this wide ablation are the following: 1) the spectral refining component is very important, emphasizing the initial intuition that preserving the object consistency in space and time using our proposed spectral approach improves the overall performance in tracking. 2) The quality of the input in our SFTrack++ method is important, but the more methods we use, the better. 3) We obtain better results using only one single spectral iteration. %Having a learning component inside the spectral iteration might have an effect of distilling the method and compensate for the distance between ground-truth and prediction before the spectral refinement into one single spectral step.
We tried the loss functions mentioned in the protocol, but we did not see relevant variations in IoU for the validation set, so we settle to BCE.

\begin{figure}[t]
	\begin{center}
		\includegraphics[width=0.99\linewidth]{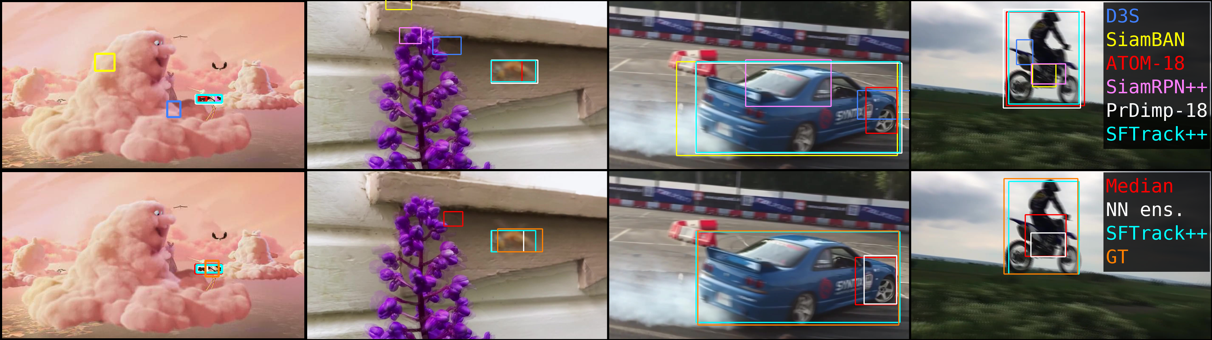}
	\end{center}
	\caption{Qualitative results. We compare in the first line SFTrack++ with the input from individual methods. In the second line, we show the ground truth (in orange) and ensemble methods results that receive the same input as SFTrack++. We notice that even though the other ensembles fail to find a good bounding box, SFTrack++ manages to combine the input methods better, even in cases with a high variance among input methods.}
	\label{fig: qual_res}
\end{figure}

\paragraph{Qualitative results.} Since SFTrack++ is an ensemble method, we show in Fig.~\ref{fig: qual_res} difficult cases and how it compares with individual methods and with other ensembles, starting from the same input (namely, all single methods from the first line, as explained in Sec.~\ref{sec: results}). We see how our method outperforms the others, even in those hard cases where an agreement seems hard to achieve.

\begingroup
\setlength{\tabcolsep}{4pt} % Default value: 6pt
\begin{table}[t]
    \begin{center}
        \begin{tabular}{l cccc}
        \toprule
        \textbf{SFTrack++} variations & \multicolumn{1}{c}{\textbf{OTB}} &
        \multicolumn{1}{c}{\textbf{UAV}} &
        \multicolumn{1}{c}{\textbf{NFS}} & \multicolumn{1}{c}{\textbf{OTB+UAV+NFS}} \\
        % \midrule
        \cmidrule(l){1-1}
        \cmidrule(l){2-5}
        w/o Spectral Refinement (phase two) & \textbf{71.6} & 60.5 & 60.9 & 64.0\\
        w/o NN$_{segm2bbox}$ (phase three) & 65.5 & 57.4 & 58.5 & 60.2\\
        \cmidrule(l){1-1}
        \cmidrule(l){2-5}
        % SFTrack++$(\text{Best method})$
        Median (over 5 methods) as input & 70.8 & 60.8 & 60.0 & 63.7\\
        Best method (PrDimp-18) as input & 67.1 & 59.7 & 61.3 & 62.5\\
        % SFTrack++$(\text{Ensemble})$ 
        Top 3 methods as input & 64.8 & 60.8 & 61.1 & 62.1\\
        \cmidrule(l){1-1}
        \cmidrule(l){2-5}
        2 spectral iterations & 70.3 & 60.9 & 61.8 & 64.1\\
        3 spectral iterations & 68.0 & 61.0 & 60.1 & 62.9\\
        \cmidrule(l){1-1}
        \cmidrule(l){2-5}
        \textbf{SFTrack++} (1 iter, 5 methods) & 70.3 & \textbf{61.2} & \textbf{62.4} & \textbf{64.5}\\
        \bottomrule
        \end{tabular}
    \end{center}
    \caption{Ablations on OTB100+UAV123+NFS30 benchmarks. In the first group we remove from the pipeline phase 2 and phase 3, respectively. The results show that both components are crucial for the method. Next, we vary the number of input methods (1 to 5) but also their quality (best or median). We see that even the quality of the input matters, using more methods as input improves the overall performance. In the last part, we validate the number of spectral iterations, a single iteration achieving the best score, which slowly degrades over more iterations.}
    \label{tab:ablations}
\end{table}
\endgroup

\section{Findings}
The pre-registered hypotheses that SFTrack++, our spectral approach that improves the space-time consistency of an object, improves the tracking performance proved to be valid.

Our method is not choosing the best input method, but it learns how to combine all inputs towards a superior performance, not only w.r.t. each input, but also w.r.t. a basic and a learned ensemble solution. SFTrack++ is robust, with a very low variance when re-training the entire pipeline from scratch, as shown in Tab.~\ref{tab:baselines}. We also noticed that the output of our method is very bold and it does not depend on a carefully chosen threshold. In Tab.~\ref{tab:ablations} we show the importance of integrating the spectral clustering module in our SFTrack++ algorithm. 

As a side observation, SFTrack++ has a weak performance on very small tracked objects when compared with single methods (UAV videos), but when compared with other ensembles, it achieves top results. This might be due to the large variance in the chosen individual methods predictions for the small objects.

In conclusion, SFTrack++ pre-registered proposal hypothesis validates through the proposed experimental protocol, with clear positive results.

\section{Documented Modifications}
\label{sec: modification}
There were several aspects where we need to deviate from the original protocol. Those aspects did not affect the core proposal and were mainly motivated by making the experiments less expensive in terms of computational cost, as detailed below.

\textbf{Number of Benchmarks and Input Trackers.} For each input tracker method considered, we run it in advance on all benchmarks (on training, valid, and test splits) to generate pre-processed input. We also generate the ground-truth bounding box segmentations for all benchmarks. This speeds up our training, making the overall training and testing self-contained,
independent w.r.t. the input methods' code. We drop out the VOT2018 benchmark because its evaluation protocol consists of running a tracker on sub-videos, therefore we should have run all the input trackers code online, and this would have been too time-consuming. We resized each frame to keep its aspect ratio, having its maximum dimension of $480$ pixels. For training, for each video in the tracking benchmarks, we used only a sample with 5 frames. The pre-processed data for one single tracker, on segmentation and tracking benchmarks, for training, valid, and testing splits takes $\approx1$ TB (without LaSOT). Since LaSOT has a very large number of frames, we decided to drop it out to make the experiment time manageable. We considered that if we test our proposal on $5$ trackers and $5$ benchmarks, our core concept of the proposal would not be affected and the results would be sufficiently general and conclusive. We chose the $5$ trackers (out of the $10$ in the proposal) that were the easiest to integrate with the PyTracking~\cite{pytracking} framework.

\textbf{$NN_{feat2segm}$ module.} Since the considered trackers were very different, we couldn't find a proper way to extract similar features from each tracker model and we decided to drop this ablation.

\textbf{Bounding box regression.} For extracting the bounding box coordinates out of the segmentation mask we use in all our experiments the region proposal from scikit-learn~\cite{scikit-learn}, with a $0.75$ threshold for binarization. We did not perform an ablation study on bounding box regression because this would not be our contribution and did not influence our core proposal, this solution for bounding box regression being good enough to emphasize what we followed in our approach.

\bibliography{sftrack}

\begin{thebibliography}{50}
\providecommand{\natexlab}[1]{#1}
\providecommand{\url}[1]{\texttt{#1}}
\expandafter\ifx\csname urlstyle\endcsname\relax
  \providecommand{\doi}[1]{doi: #1}\else
  \providecommand{\doi}{doi: \begingroup \urlstyle{rm}\Url}\fi

\bibitem[Abraham and Khan(2019)]{focal-tversky}
Nabila Abraham and Naimul~Mefraz Khan.
\newblock A novel focal tversky loss function with improved attention u-net for
  lesion segmentation.
\newblock In \emph{{ISBI}}, 2019.

\bibitem[Bertinetto et~al.(2016{\natexlab{a}})Bertinetto, Henriques, Valmadre,
  Torr, and Vedaldi]{oneshot-tracker}
Luca Bertinetto, Jo{\~{a}}o~F. Henriques, Jack Valmadre, Philip H.~S. Torr, and
  Andrea Vedaldi.
\newblock Learning feed-forward one-shot learners.
\newblock In \emph{NIPS}, 2016{\natexlab{a}}.

\bibitem[Bertinetto et~al.(2016{\natexlab{b}})Bertinetto, Valmadre, Henriques,
  Vedaldi, and Torr]{siamfc}
Luca Bertinetto, Jack Valmadre, Jo{\~{a}}o~F. Henriques, Andrea Vedaldi, and
  Philip H.~S. Torr.
\newblock Fully-convolutional siamese networks for object tracking.
\newblock In Gang Hua and Herv{\'{e}} J{\'{e}}gou, editors, \emph{{ECCV}
  Workshops}, 2016{\natexlab{b}}.

\bibitem[Bhat et~al.(2020)Bhat, Danelljan, Gool, and Timofte]{kys}
Goutam Bhat, Martin Danelljan, Luc~Van Gool, and Radu Timofte.
\newblock Know your surroundings: Exploiting scene information for object
  tracking.
\newblock In \emph{{ECCV}}, 2020.

\bibitem[Bradski and Kaehler(2008)]{opencv}
Gary Bradski and Adrian Kaehler.
\newblock \emph{Learning OpenCV: Computer vision with the OpenCV library}.
\newblock " O'Reilly Media, Inc.", 2008.

\bibitem[Burceanu and Leordeanu(2018)]{multi-path-tracking}
Elena Burceanu and Marius Leordeanu.
\newblock Learning a robust society of tracking parts using co-occurrence
  constraints.
\newblock In \emph{{ECCV} Workshops}, 2018.

\bibitem[Burceanu and Leordeanu(2020)]{sfseg}
Elena Burceanu and Marius Leordeanu.
\newblock A 3d convolutional approach to spectral object segmentation in space
  and time.
\newblock In Christian Bessiere, editor, \emph{Proceedings of the Twenty-Ninth
  International Joint Conference on Artificial Intelligence, {IJCAI} 2020},
  pages 495--501. ijcai.org, 2020.
\newblock \doi{10.24963 / ijcai.2020 / 69}.
\newblock URL \url{https://doi.org/10.24963/ijcai.2020/69}.

\bibitem[Caelles et~al.(2017)Caelles, Maninis, Pont-Tuset, Leal-Taix{\'e},
  Cremers, and {Van Gool}]{osvos}
S.~Caelles, K.~Maninis, J.~Pont-Tuset, L.~Leal-Taix{\'e}, D.~Cremers, and
  L.~{Van Gool}.
\newblock One-shot video object segmentation.
\newblock \emph{CVPR}, 2017.

\bibitem[Chen et~al.(2020)Chen, Zhong, Li, Zhang, and Ji]{siamban}
Zedu Chen, Bineng Zhong, Guorong Li, Shengping Zhang, and Rongrong Ji.
\newblock Siamese box adaptive network for visual tracking.
\newblock In \emph{{CVPR}}, 2020.

\bibitem[Dai et~al.(2020)Dai, Zhang, Wang, Li, Lu, and Yang]{ltmu}
Kenan Dai, Yunhua Zhang, Dong Wang, Jianhua Li, Huchuan Lu, and Xiaoyun Yang.
\newblock High-performance long-term tracking with meta-updater.
\newblock In \emph{{CVPR}}, 2020.

\bibitem[Danelljan et~al.(2017)Danelljan, Bhat, Khan, and Felsberg]{eco}
Martin Danelljan, Goutam Bhat, Fahad~Shahbaz Khan, and Michael Felsberg.
\newblock {ECO:} efficient convolution operators for tracking.
\newblock In \emph{{CVPR}}, 2017.

\bibitem[Danelljan et~al.(2019)Danelljan, Bhat, Khan, and Felsberg]{atom}
Martin Danelljan, Goutam Bhat, Fahad~Shahbaz Khan, and Michael Felsberg.
\newblock {ATOM:} accurate tracking by overlap maximization.
\newblock In \emph{{CVPR}}, 2019.

\bibitem[Danelljan et~al.(2020)Danelljan, Gool, and Timofte]{prdimp}
Martin Danelljan, Luc~Van Gool, and Radu Timofte.
\newblock Probabilistic regression for visual tracking.
\newblock In \emph{{CVPR}}, 2020.

\bibitem[Ding et~al.(2001)Ding, He, Zha, Gu, and Simon]{cut_min_max}
Chris H.~Q. Ding, Xiaofeng He, Hongyuan Zha, Ming Gu, and Horst~D. Simon.
\newblock A min-max cut algorithm for graph partitioning and data clustering.
\newblock In \emph{ICDM}, 2001.

\bibitem[Fan et~al.(2019)Fan, Lin, Yang, Chu, Deng, Yu, Bai, Xu, Liao, and
  Ling]{lasot}
Heng Fan, Liting Lin, Fan Yang, Peng Chu, Ge~Deng, Sijia Yu, Hexin Bai, Yong
  Xu, Chunyuan Liao, and Haibin Ling.
\newblock Lasot: {A} high-quality benchmark for large-scale single object
  tracking.
\newblock In \emph{{CVPR}}, 2019.

\bibitem[Galoogahi et~al.(2017)Galoogahi, Fagg, Huang, Ramanan, and Lucey]{nfs}
Hamed~Kiani Galoogahi, Ashton Fagg, Chen Huang, Deva Ramanan, and Simon Lucey.
\newblock Need for speed: {A} benchmark for higher frame rate object tracking.
\newblock In \emph{{ICCV}}, 2017.

\bibitem[Huang et~al.(2019)Huang, Zhao, and Huang]{got-10k}
Lianghua Huang, Xin Zhao, and Kaiqi Huang.
\newblock Got-10k: A large high-diversity benchmark for generic object tracking
  in the wild.
\newblock \emph{IEEE Transactions on Pattern Analysis and Machine
  Intelligence}, 2019.

\bibitem[Jabri et~al.(2020{\natexlab{a}})Jabri, Owens, and
  Efros]{contrastive-random-walk}
Allan Jabri, Andrew Owens, and Alexei~A. Efros.
\newblock Space-time correspondence as a contrastive random walk.
\newblock \emph{CVPR}, 2020{\natexlab{a}}.

\bibitem[Jabri et~al.(2020{\natexlab{b}})Jabri, Owens, and
  Efros]{cycle-random-walk}
Allan Jabri, Andrew Owens, and Alexei~A. Efros.
\newblock Space-time correspondence as a contrastive random walk.
\newblock \emph{CoRR}, 2020{\natexlab{b}}.

\bibitem[Kristan et~al.(2018)Kristan, Leonardis, Matas, Felsberg, Pflugfelder,
  Zajc, Voj{\'{\i}}r, Bhat, Lukezic, Eldesokey, Fern{\'{a}}ndez, and
  et~al.]{vot-18}
Matej Kristan, Ales Leonardis, Jiri Matas, Michael Felsberg, Roman~P.
  Pflugfelder, Luka~Cehovin Zajc, Tom{\'{a}}s Voj{\'{\i}}r, Goutam Bhat, Alan
  Lukezic, Abdelrahman Eldesokey, Gustavo Fern{\'{a}}ndez, and et~al.
\newblock The sixth visual object tracking {VOT2018} challenge results.
\newblock In \emph{{ECCV} 2018 Workshops}, 2018.

\bibitem[Leordeanu and Hebert(2005)]{marius_iccv2005}
Marius Leordeanu and Martial Hebert.
\newblock A spectral technique for correspondence problems using pairwise
  constraints.
\newblock \emph{{ICCV}}, 2005.

\bibitem[Li et~al.(2018)Li, Yan, Wu, Zhu, and Hu]{siam-rpn}
Bo~Li, Junjie Yan, Wei Wu, Zheng Zhu, and Xiaolin Hu.
\newblock High performance visual tracking with siamese region proposal
  network.
\newblock In \emph{{CVPR}}, 2018.

\bibitem[Li et~al.(2019)Li, Wu, Wang, Zhang, Xing, and Yan]{siamrpn++}
Bo~Li, Wei Wu, Qiang Wang, Fangyi Zhang, Junliang Xing, and Junjie Yan.
\newblock Siamrpn++: Evolution of siamese visual tracking with very deep
  networks.
\newblock In \emph{{CVPR}}, 2019.

\bibitem[Loshchilov and Hutter(2019)]{adamw-amsgrad}
Ilya Loshchilov and Frank Hutter.
\newblock Decoupled weight decay regularization.
\newblock In \emph{{ICLR}}, 2019.

\bibitem[Lukezic et~al.(2020)Lukezic, Matas, and Kristan]{d3s}
Alan Lukezic, Jiri Matas, and Matej Kristan.
\newblock {D3S} - {A} discriminative single shot segmentation tracker.
\newblock In \emph{{CVPR}}, 2020.

\bibitem[Martin~Danelljan(2019)]{pytracking}
Christoph~Mayer Martin~Danelljan, Goutam~Bhat.
\newblock {PyTracking}, 2019.
\newblock URL \url{"https://github.com/visionml/pytracking"}.

\bibitem[Meila and Shi(2001)]{meila_shi}
Marina Meila and Jianbo Shi.
\newblock A random walks view of spectral segmentation.
\newblock In \emph{AISTATS}, 2001.

\bibitem[Mueller et~al.(2016)Mueller, Smith, and Ghanem]{uav123}
Matthias Mueller, Neil Smith, and Bernard Ghanem.
\newblock A benchmark and simulator for {UAV} tracking.
\newblock In Bastian Leibe, Jiri Matas, Nicu Sebe, and Max Welling, editors,
  \emph{{ECCV}}, 2016.

\bibitem[M{\"{u}}ller et~al.(2018)M{\"{u}}ller, Bibi, Giancola, Al{-}Subaihi,
  and Ghanem]{tracking-net}
Matthias M{\"{u}}ller, Adel Bibi, Silvio Giancola, Salman Al{-}Subaihi, and
  Bernard Ghanem.
\newblock Trackingnet: {A} large-scale dataset and benchmark for object
  tracking in the wild.
\newblock In \emph{{ECCV}}, 2018.

\bibitem[Ng et~al.(2001)Ng, Jordan, and Weiss]{NJW}
Andrew~Y. Ng, Michael~I. Jordan, and Yair Weiss.
\newblock On spectral clustering: Analysis and an algorithm.
\newblock In \emph{{NIPS}}, 2001.

\bibitem[Park and Berg(2018)]{meta-tracker}
Eunbyung Park and Alexander~C. Berg.
\newblock Meta-tracker: Fast and robust online adaptation for visual object
  trackers.
\newblock In \emph{{ECCV}}, 2018.

\bibitem[Pedregosa et~al.(2011)Pedregosa, Varoquaux, Gramfort, Michel, Thirion,
  Grisel, Blondel, Prettenhofer, Weiss, Dubourg, Vanderplas, Passos,
  Cournapeau, Brucher, Perrot, and Duchesnay]{scikit-learn}
F.~Pedregosa, G.~Varoquaux, A.~Gramfort, V.~Michel, B.~Thirion, O.~Grisel,
  M.~Blondel, P.~Prettenhofer, R.~Weiss, V.~Dubourg, J.~Vanderplas, A.~Passos,
  D.~Cournapeau, M.~Brucher, M.~Perrot, and E.~Duchesnay.
\newblock Scikit-learn: Machine learning in {P}ython.
\newblock \emph{Journal of Machine Learning Research}, 2011.

\bibitem[Perazzi et~al.(2016)Perazzi, Pont-Tuset, McWilliams, {Van Gool},
  Gross, and Sorkine-Hornung]{davis-2016}
F.~Perazzi, J.~Pont-Tuset, B.~McWilliams, L.~{Van Gool}, M.~Gross, and
  A.~Sorkine-Hornung.
\newblock A benchmark dataset and evaluation methodology for video object
  segmentation.
\newblock In \emph{{CVPR}}, 2016.

\bibitem[Pont-Tuset et~al.(2017)Pont-Tuset, Perazzi, Caelles, Arbel\'aez,
  Sorkine-Hornung, and {Van Gool}]{davis-2017}
Jordi Pont-Tuset, Federico Perazzi, Sergi Caelles, Pablo Arbel\'aez, Alexander
  Sorkine-Hornung, and Luc {Van Gool}.
\newblock The 2017 davis challenge on video object segmentation.
\newblock \emph{arXiv:1704.00675}, 2017.

\bibitem[Ronneberger et~al.(2015)Ronneberger, Fischer, and Brox]{unet}
Olaf Ronneberger, Philipp Fischer, and Thomas Brox.
\newblock U-net: Convolutional networks for biomedical image segmentation.
\newblock In \emph{{MICCAI}}, 2015.

\bibitem[Sarkar and Soundararajan(2000)]{cut_average}
Sudeep Sarkar and Padmanabhan Soundararajan.
\newblock Supervised learning of large perceptual organization: Graph spectral
  partitioning and learning automata.
\newblock \emph{{PAMI}}, 2000.

\bibitem[Shi and Malik(2000)]{img_normalized_cut_malik_2000}
Jianbo Shi and Jitendra Malik.
\newblock Normalized cuts and image segmentation.
\newblock In \emph{{PAMI}}, 2000.

\bibitem[Sudre et~al.(2017)Sudre, Li, Vercauteren, Ourselin, and
  Cardoso]{wdice-loss}
Carole~H. Sudre, Wenqi Li, Tom Vercauteren, S{\'{e}}bastien Ourselin, and
  M.~Jorge Cardoso.
\newblock Generalised dice overlap as a deep learning loss function for highly
  unbalanced segmentations.
\newblock In \emph{{MICCAI}}, 2017.

\bibitem[Tao et~al.(2016)Tao, Gavves, and Smeulders]{siam-inst-search}
Ran Tao, Efstratios Gavves, and Arnold W.~M. Smeulders.
\newblock Siamese instance search for tracking.
\newblock In \emph{{CVPR}}, 2016.

\bibitem[Torsello et~al.(2006)Torsello, Bul{\`{o}}, and
  Pelillo]{directed-affinities}
Andrea Torsello, Samuel~Rota Bul{\`{o}}, and Marcello Pelillo.
\newblock Grouping with asymmetric affinities: {A} game-theoretic perspective.
\newblock In \emph{CVPR}, 2006.

\bibitem[Voigtlaender et~al.(2019)Voigtlaender, Krause, Osep, Luiten, Sekar,
  Geiger, and Leibe]{mots}
Paul Voigtlaender, Michael Krause, Aljosa Osep, Jonathon Luiten, Berin
  Balachandar~Gnana Sekar, Andreas Geiger, and Bastian Leibe.
\newblock {MOTS:} multi-object tracking and segmentation.
\newblock In \emph{{CVPR}}, 2019.

\bibitem[Voigtlaender et~al.(2020)Voigtlaender, Luiten, Torr, and
  Leibe]{siam-rcnn}
Paul Voigtlaender, Jonathon Luiten, Philip H.~S. Torr, and Bastian Leibe.
\newblock Siam {R-CNN:} visual tracking by re-detection.
\newblock In \emph{{CVPR}}, 2020.

\bibitem[Wang et~al.(2020)Wang, Luo, Sun, Xiong, and
  Zeng]{meta-learning-tracker}
Guangting Wang, Chong Luo, Xiaoyan Sun, Zhiwei Xiong, and Wenjun Zeng.
\newblock Tracking by instance detection: {A} meta-learning approach.
\newblock In \emph{{CVPR}}, 2020.

\bibitem[Wang and Chung(2018)]{focal-dice}
Pei Wang and Albert C.~S. Chung.
\newblock Focal dice loss and image dilation for brain tumor segmentation.
\newblock In Danail Stoyanov, Zeike Taylor, Gustavo Carneiro, Tanveer~F.
  Syeda{-}Mahmood, and et~al., editors, \emph{{MICCAI}}, 2018.

\bibitem[Wang et~al.(2019)Wang, Zhang, Bertinetto, Hu, and Torr]{segtrack}
Qiang Wang, Li~Zhang, Luca Bertinetto, Weiming Hu, and Philip H.~S. Torr.
\newblock Fast online object tracking and segmentation: {A} unifying approach.
\newblock In \emph{{CVPR}}, 2019.

\bibitem[Wu et~al.(2015)Wu, Lim, and Yang]{otb100}
Yi~Wu, Jongwoo Lim, and Ming{-}Hsuan Yang.
\newblock Object tracking benchmark.
\newblock \emph{{IEEE} Trans. Pattern Anal. Mach. Intell.}, 2015.

\bibitem[Xu et~al.(2020)Xu, Wang, Li, Yuan, and Yu]{siamfc++}
Yinda Xu, Zeyu Wang, Zuoxin Li, Ye~Yuan, and Gang Yu.
\newblock Siamfc++: Towards robust and accurate visual tracking with target
  estimation guidelines.
\newblock In \emph{{AAAI}}, 2020.

\bibitem[Yang et~al.(2019)Yang, Fan, and Xu]{Youtube-VIS}
Linjie Yang, Yuchen Fan, and Ning Xu.
\newblock Video instance segmentation.
\newblock In \emph{{ICCV}}, 2019.

\bibitem[Yu and Shi(2001)]{directed-groups}
Stella~X. Yu and Jianbo Shi.
\newblock Grouping with directed relationships.
\newblock In \emph{EMMCVPR 2001}, 2001.

\bibitem[Zhang and Peng(2020)]{ocean}
Zhipeng Zhang and Houwen Peng.
\newblock Ocean: Object-aware anchor-free tracking.
\newblock In \emph{{ECCV}}, 2020.

\end{thebibliography}

\end{document}